\documentclass[11pt]{article}
\usepackage{coling2020}
\usepackage{times}
\usepackage{url}
\usepackage{latexsym}
\usepackage{epigraph}
\usepackage{booktabs}
\usepackage{xcolor}
\usepackage{hyperref}
\usepackage{csquotes}

\usepackage{graphicx}
\usepackage{tikz}
\usepackage{gb4e}

\begin{document}









\begin{minipage}[c]{\textwidth}
    \centering
    \begin{tikzpicture}[remember picture, overlay]
        \node [inner sep = 0] at (current page.center) {\includegraphics[width=\paperwidth]{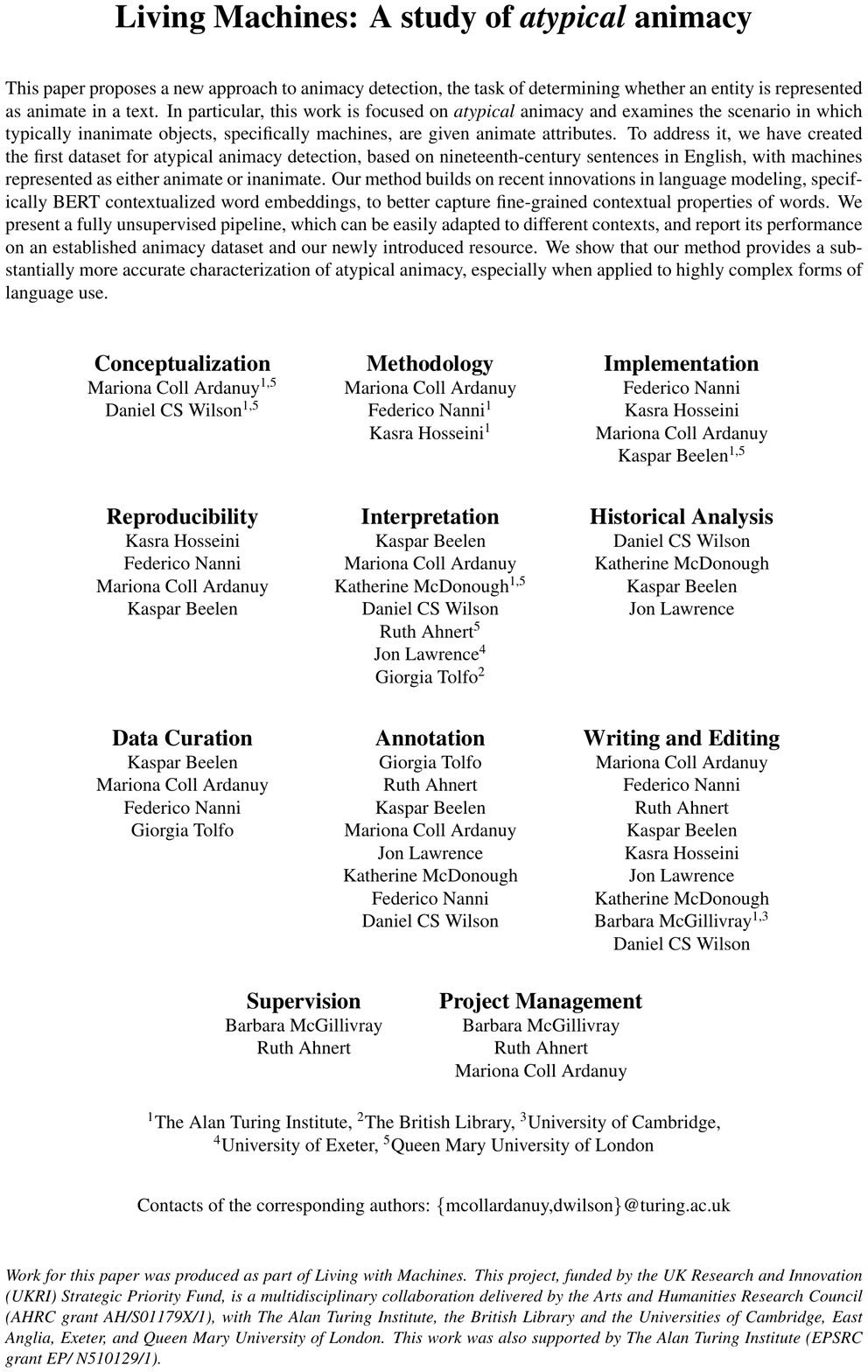}};
    \end{tikzpicture}
\end{minipage}

\newpage


\section{Introduction} \label{intro}
Animacy is the property of being alive. Although the \textit{perception} of a given entity as animate (or not) tends to align correctly with its biological animacy, discrepancies are not uncommon. These may arise from the ways different entities are unconsciously perceived, or from the deliberately figurative use of animate expressions to describe inanimate entities (or vice versa). Machines sit at the fuzzy boundary between animacy and inanimacy \cite{turing1950computing,yamamoto1999animacy}. In this paper, we provide a basis for examining how machines were imagined during the nineteenth century as everything from lifeless mechanical objects to living beings, or even human-like agents that feel, think, and love. We focus on texts from nineteenth-century Britain, a society being transformed by industrialization, as a good candidate for studying the broader issue.

This paper applies state-of-the-art contextualized word representations, trained using the BERT architecture \cite{devlin-etal-2019-bert}, to animacy detection. In contrast to previous research, this paper provides an in-depth exploration of the ambiguities and figurative aspects that characterize animacy in natural language, and analyzes how context shapes animacy. Context is constitutive of meaning \cite[3.3]{wittgenstein2012tractatus}, an observation acknowledged by generations of scholars, but which is still difficult to apply to its full extent in computational models of language. We show how the increased sensitivity of BERT-based models to contextual cues can be exploited to analyze how the same entity (e.g., a machine) can be at once represented as animate or inanimate depending on the intent of the writer.

This paper makes several contributions: we present an unsupervised method to detect animacy that is highly sensitive to the context and therefore suited to capture not only typical animacy, but especially atypical animacy. We provide the first benchmark for atypical animacy detection based on a dataset of nineteenth-century sentences in English with machines represented as animate and inanimate. We conduct an extensive quantitative evaluation of our approach in comparison with strong baselines on both an established animacy dataset and on our newly introduced resource, and demonstrate the generalizability of our approach. Finally, we discuss the distinction between animacy and humanness which, we believe, has the potential to shed light on the related (but separate) historical process of dehumanization through the language of mechanization, but which lies beyond the scope of the present paper.  

Atypical observations are rare by definition. Because of this, addressing them is often an ungratifying undertaking, as they can only marginally improve the accuracy of general natural language processing systems on existing benchmarks, if at all. And yet, precisely because of this, atypical observations tend to acquire a certain salience from a qualitative and interpretative point of view. For the humanities scholar and the linguist, such deviations prove particularly interesting objects of study because they flout expectations.

\section{Related work}
Animacy and its relation to (human) cognition has been extensively studied in a range of linguistic fields, from neurolinguistics and language acquisition research \cite{opfer,gao} to morphology and syntax \cite{rosenbach,mclaughlin,vihman2019effects}. There is evidence that animacy is not a fixed property of lexical items but is subject to their context \cite{mante}. This points to a more nuanced and graduated view of animacy than a binary distinction between ``animate'' and ``inanimate'' \cite{peltola2018interspecies,bayanati2019humans}, which results in a hierarchy of entities that reflects notions of agency, closeness to the speaker, individuation, and empathy \cite{comrie,yamamoto1999animacy,croft2002typology}. \newcite{yamamoto1999animacy} identifies modern machines as one of the most prominent examples at the frontier area between animacy and inanimacy.

The distinction between animate and inanimate is a fundamental aspect of cognition and language, and has been shown to be a useful feature in natural language processing (NLP), in tasks such as co-reference and anaphora resolution \cite{orasan2007np,poesio2008anaphoric,raghunathan2010multi,lee}, word sense disambiguation \cite{chen2006empirical,ovrelid}, and semantic role labeling \cite{connor2010starting}. Earlier approaches to animacy detection relied on semantic lexicons (such as WordNet, \newcite{fellbaum1998}) combined with syntactic analysis \cite{evans}, or developed machine-learning classifiers that use syntactic and morphological features \cite{ovrelid}. More recently, \newcite{karsdorp2015animacy} focused on Dutch folk tales and trained a classifier to identify animate entities based on a combination of linguistic features and word embeddings trained using a skip-gram model. They showed that close-to-optimal scores could be achieved using word embeddings alone. \newcite{jahan2018new} developed a hybrid classification system which relies on rich linguistic text processing, by combining static word embeddings with a number of hand-built rules 
to compute the animacy of referring expressions and co-reference chains. Previous research \cite{karsdorp2015animacy,jahan2018new} has acknowledged the importance of context in atypical animacy, but it has not explicitly tackled it, or attempted to quantify how well existing methods have handled such complexities.

Whereas static word representations such as Word2Vec \cite{mikolov2013efficient} have been shown to perform well in typical animacy detection tasks, we argue that they are not capable of detecting atypical cases of animacy, as by definition 
this must arise from the context, and not the target entity itself. The emergence of contextualized word representations has yielded significant advances in many NLP tasks \cite{peters2017semi,radford2018improving,devlin-etal-2019-bert}. Unlike their static counterparts, they are optimized to capture the representations of target words in their contexts, and are 
more sensitive to context-dependent aspects of meaning. BERT (Bidirectional Encoder Representations from Transformers, \newcite{devlin-etal-2019-bert}) incorporates the latest improvements in language modeling. Through its deep bidirectionality and 
self-attention mechanism, it has become one of the most successful attempts to train context-sensitive language models. BERT is pre-trained on two tasks: Masked Language Model (MLM), which tries to predict a masked token based on 
its left and right context, and Next Sentence Prediction (NSP), which tries to predict the following sentence through a binary classification task. This dual learning objective ensures that the contextual representations of words are learned, also across sentences. Its simple and efficient fine-tuning mechanism allows BERT to easily adapt to different tasks and domains.

\section{Method}\label{sec:method}

In this section, we describe our approach to determine the animacy of a target expression in its context. The intuition behind our method is the following: an entity becomes animate in a given context if it occurs in a position in which one would typically expect a living entity. More specifically, given a sentence in which a target expression has been adequately masked, we rely on contextualized masked language models to provide ranked predictions for the masked element, as shown in example \ref{ex:mask2bert}:

\begin{exe}
	\ex \label{ex:mask2bert} \textbf{Original sentence:} And why should one say that the machine does not live?\\
	\textbf{Masked sentence:} And why should one say that the [MASK] does not live?\\
	\textbf{Predictions with scores:} \textit{man} ($5.0788$), \textit{person} ($4.4484$), \textit{other} ($4.1866$), \textit{child} ($4.1600$), \textit{king} ($4.1510$), \textit{patient} ($4.1249$), \textit{one} ($4.1141$), \textit{stranger} ($4.1067$), ...
\end{exe}

\noindent We then determine the animacy of the masked expression\footnote{Note that this can be a multi-word expression.} by averaging the animacy of the top $\kappa$ tokens that have been predicted to fill the mask in the sentence. While this may initially seem to be a circular argument, the fundamentally probabilistic nature of language models means that we are in fact replacing the masked element with tokens that have a high probability of occurring in this context. Our method rests on the assumption that, given a context requiring an animate entity, a contextualized language model should predict tokens corresponding to conventionally animate entities. 



We use a BERT language model to predict a number of possible fillers given a sentence with a masked expression and their corresponding probability scores. We then use WordNet,\footnote{\href{https://wordnet.princeton.edu/}{https://wordnet.princeton.edu/}.} a lexical database that encodes relations between word senses \cite{fellbaum1998}, to determine whether the predicted tokens correspond to typically animate or inanimate entities. Tokens can be ambiguous: the same token can be used for several different word senses, some of which may correspond to living entities and some others not. For example, the word `dresser' has several meanings, including a profession -- typically animate -- and a piece of furniture -- typically inanimate. We disambiguate each predicted token to its most relevant sense in WordNet by measuring the similarity between the original sentence and the gloss of each WordNet sense. Inspired by previous research on distributional semantic models for word sense disambiguation \cite{basile2014enhanced}, we use a BERT-adapted version of the Lesk algorithm, which adopts recent advancements in transformer-based sentence representations \cite{reimers2019sentence}.

WordNet organizes nouns according to hierarchies, which eventually converge at the root node \textit{entity}. Senses of nouns that correspond to living entities fall under the \textit{living\_thing} node, which is the common parent of the person, animal, plant, and microorganism classes, among others. Therefore, we determine whether each predicted token corresponds to an animate or inanimate entity based on whether its disambiguated sense is a descendant of the \textit{living\_thing} node.\footnote{Pronouns are not represented in WordNet. Because of this, we also automatically assign animacy to a predicted token if it is included in a closed list of pronouns typically referring to persons, such as `she', `him', and `yourselves'.} Finally, we produce a single animacy score between 0 and 1 for the masked element, by averaging the animacy values (i.e.~0 if inanimate, 1 if animate) of the predicted tokens, weighted by their probability score. We find the optimal animacy threshold $\tau$ and cutoff $\kappa$ (i.e.~number of predicted tokens) through experimentation.\footnote{Code and experimental results are available at \href{https://github.com/Living-with-machines/AtypicalAnimacy}{https://github.com/Living-with-machines/AtypicalAnimacy}.}

\section{Data}
We use two datasets to evaluate the performance of our algorithm: the first is derived from the data released by \newcite{jahan2018new}, while the second has been created by us for specifically testing detection of unconventional animacy, in nineteenth-century English texts in particular. Both datasets are described in sections \ref{corpus:jahan2108} and \ref{corpus:19thC} respectively, and summarized and discussed in section \ref{corpus:summary}.

\subsection{The \textit{Stories} animacy dataset}\label{corpus:jahan2108}
In their paper, \newcite{jahan2018new} used a collection of English-translated stories that already contained several layers of linguistic annotation \cite{finlayson2014n2,finlayson2017propplearner}. The authors enriched the texts with animacy annotations at the level of co-reference chains and of their referring expressions. They reported a near-perfect inter-annotation agreement (Cohen's $\kappa=0.99$). We restructured their data to make it compatible with our approach, which takes full sentences as inputs. This process resulted in a new dataset (henceforth \textbf{\textit{Stories}} dataset) consisting of 18,708 sentences, each of which contains a target expression annotated with animacy (see some examples in Table \ref{tab:jahanExamples}).\footnote{Unfortunately, we were not able to reproduce the same number of target expressions as are reported in \newcite{jahan2018new}, but we provide the code used to generate our datasets from the original corpus. See Table \ref{tab:jahanVs19thC} for a summary of the \textit{Stories} dataset.}

\begin{table*}[h!]
\small
\begin{tabular}{lp{0.64\linewidth}c}
\toprule
Target expression & Sentence & Animacy \\
\midrule
noise of the leaves & ``No it was only the \textbf{noise of the leaves}.'' & 0 \\
need & He whistled and shouted with a mighty hiss, in a truly powerful voice: ``Magic steed, horse of my \textbf{need}! & 0 \\
long time or a short time & Nikita fought him for a \textbf{long time or a short time}; in any event, he defeated him. & 0 \\
merchant's daughter & The merchant's daughter took the letter, burst into tears, and prepared to go, taking also her maidservant; and no one could distinguish this maid from the \textbf{merchant's daughter}, they were so like each other. & 1 \\
twelve-headed dragon & They began to fight, and after a long struggle or a short struggle, Prince Ivan defeated the \textbf{twelve-headed dragon}, slashed open his trunk, and found the coffer in his right side. & 1 \\
She & \textbf{She} wept, but still could not awaken Prince Ivan. & 1 \\
\bottomrule
\end{tabular}
\caption{Examples of sentences from the \textit{Stories} dataset, with the target expressions and their animacy values, derived from the work by Jahan et al.~(2018).}
\label{tab:jahanExamples}
\end{table*}

\subsection{The \textit{19thC Machines} animacy dataset}\label{corpus:19thC}
The \textit{Stories} dataset is largely composed of target expressions that correspond to either typically animate or typically inanimate entities. Even though some cases of unconventional animacy can be found (folktales, in particular, are richer in typically inanimate entities that become animate), these account for a very small proportion of the data.\footnote{Note that this is based on observation, as the number of atypical cases is not provided in \newcite{jahan2018new}.} We decided to create our own dataset (henceforth \textbf{\textit{19thC Machines}} dataset) to gain a better sense of the suitability of our method to the problem of atypical animacy detection, with particular attention to the case of animacy of machines in nineteenth-century texts.


We extracted sentences containing nouns that correspond to types of machines from an open dataset of nineteenth-century books (from now on \textit{19thC BL Books}).\footnote{This nineteenth-century book dataset was digitized by the British Library. It contains $\approx$48,200 volumes with $\approx$4.9B tokens. Data available via \href{https://data.bl.uk/digbks/db14.html}{https://data.bl.uk/digbks/db14.html}, DOI \href{https://doi.org/10.21250/db14}{https://doi.org/10.21250/db14} (British Library Labs, 2014).} Even though the OCR quality is relatively good, some noise can still be found in the dataset. In order to extract sentences which contain machine-related words, we manually selected words that occurred close to the combined vector of `machine' and `machines' in Word2vec models trained on BL books from before and after 1850 (to make sure the selection is not biased towards a particular half of the nineteenth century). We refined this list in multiple iterations, adding new words and recomputing the combined vector. The result was a stable list of generic words referring to machines across the period under investigation.\footnote{The curated words are: `machine', `machines', `machinery', `engines', `engine', `locomotive', `locomotives', `turbine', `turbines', `boiler', `boilers', `dynamo', `dynamos', `motor', `motors', `apparatus', `apparatuses', `accumulator', `accumulators', `compressor', and `compressors'. We are preparing a publication on the lexicon expansion procedure that will allow us to assess its impact on downstream tasks.} In most sentences, machines are treated as inanimate objects. We therefore employed a pooled strategy\footnote{Pooling  is an established evaluation strategy for information retrieval systems \cite{spark1975report,buckley2007bias}.} to identify meaningful sentences for annotation: we specified four animacy bands (0.0-0.25, 0.25-0.50, 0.50-0.75, and 0.75-1.00) and we used the different methods described in section \ref{sec:experiments} to obtain a fixed number of sentences for each band. This way, we obtained a large pool of sentences capturing a variety of different types of animate and inanimate contexts present in the corpus.

\subsubsection{Preliminary annotations}
For human annotators, even history and literature experts, language subtleties made this task extremely subjective. In order to gain a better understanding of the problem, we started with two preliminary annotation tasks. A first set of 100 sentences derived from the pooling process was distributed among six annotators.\footnote{A combination of computational linguists, historians, literary scholars, and data scientists.} In the first task, we masked the target word (i.e.~the machine) in each sentence and asked the annotator to fill the slot with the most likely entity between `human', `horse', and `machine', representing three levels in the animacy hierarchy: human, animal, and object \cite[185]{comrie}. We asked annotators to stick to the most literal meaning and avoid metaphorical interpretations when possible. Interestingly, even though the original masked expressions contained only instances of the lemma `machine', the annotators selected `machine' as the most likely option in only 62\% of the total number of annotations. However, the agreement was low, with a Krippendorff $\alpha$ of 0.32. This indicates that, at least in some contexts, machines seem to be interchangeable with humans and animals, and that annotators may disagree about when one is preferred over the other, especially in a historical context.

The second task was more straightforwardly related to determining the animacy of the target entity, given the same 100 sentences. We asked annotators to provide a score between -2 and 2, with -2 being definitely inanimate, -1 possibly inanimate, 1 possibly animate, and 2 definitely animate. Neutral judgements were not allowed. The agreement for this second task was low as well (Krippendorff $\alpha$ of 0.43). Neither collating the annotations into positive and negative groups nor collating slightly animate and slightly inanimate together improved inter-annotation agreement significantly. We explored the cases in which annotators disagreed, and found that the same sentence would often be annotated as highly animate by one annotator and as highly inanimate by another. This was especially true for sentences containing similes or metaphors that liken machines to humans, animals, or systems.

These preliminary annotations helped us to understand the data and improve our experimental design. Annotators were asked to leave comments and provide feedback, and agreed that both tasks were more challenging than expected, mostly due to the high incidence of figurative language, as in the sentences in example \ref{ex:humanAsMachines}.

\begin{exe}
    \item \label{ex:humanAsMachines}
    (a) \textit{He is himself but a mere \textbf{machine}, unconscious of the operations of his own mind.}\\
    (b) \textit{Our servants, like mere \textbf{machines}, move on their mercenary track without feeling.}\\
    (c) \textit{My companions treated me as a \textbf{machine}, and never in any way repaid my services.}\\
    (d) \textit{A master who looks upon thy kind, not as mere \textbf{machines}, but as valued friends.}
\end{exe}

\noindent These kinds of sentences present a very particular type of interpretative openness. In each case a human (or group of humans) is likened to a machine to suggest that their agency or animacy has been reduced. Some annotators detected an implied inanimacy of the machine, which had the rhetorical effect of rendering the humans inanimate. Conversely, for others, the comparison conjured a kind of automaton, a human-machine-hybrid, and therefore a machine made animate through its association with the human.

\subsubsection{Final annotations}\label{sec:19thAnnotations}
A subgroup of five annotators collaboratively wrote the guidelines based on their experience annotating the first batch of sentences, taking into account common discrepancies. 
After discussion, it was decided that a machine would be tagged as animate if it is described as having traits distinctive of biologically animate beings or human-specific skills, or portrayed as having feelings, emotions, or a soul. Sentences like the ones in example \ref{ex:humanAsMachines} would be considered animate, but an additional annotation layer would be provided to capture the notion of \textbf{humanness}, which would be true if the machine is portrayed as sentient and capable of specifically human emotions, and false if it used to suggest some degree of dehumanization.\footnote{For us, `animacy' encompasses `humanness': all sentences that are tagged as cases of humanness are also annotated as animate, but not the other way around. The term `humanness' is used differently in other research fields, for instance when evaluating the performance of a chatbot \cite{svenningsson2019artificial}.} The focus of this paper is on detecting atypical animacy, and we plan to further study the phenomenon of humanness (or lack thereof) in animate objects in future work.

A new batch of 600 unseen sentences was sent to the same annotators, who were asked to document their decisions, especially the ones they were most uncertain of. A subset of 100 sentences were annotated by five domain experts, for which the Krippendorff $\alpha$ was of 0.74 for animacy and 0.50 for humanness. The remaining annotations were produced by subgroups of two to three annotators. The gold standard was produced by one of the annotators and author of the guidelines, who assigned the final labels by adjudication, taking into account agreements and disagreements between annotators and their comments. We provide examples of annotations in table \ref{tab:19thCexamples}.

\begin{table*}[h!]
\small
\centering
\begin{tabular}{lp{0.64\linewidth}cc}
\toprule
Target & Sentence & Animacy & Humanness \\
\midrule
engine & In December, the first steam fire \textbf{engine} was received, and tried on the shore of Lake Monona, with one thousand feet of hose. & 0 & 0 \\
engine & It was not necessary for Jakie to slow down in order to allow the wild \textbf{engine} to come up with him; she was coming up at every revolution of her wheels. & 1 & 1 \\
locomotive & Nearly a generation had been strangely neglected to grow up un-Americanized, and the private adventurer and the \textbf{locomotive} were the untechnical missionaries to open a way for the common school. & 1 & 1 \\
machine & The worst of it was, the people were surly; not one would get out of our way until the last minute, and many pretended not to see us coming, though the \textbf{machine}, held in by the brake, squeaked a pitiful warning. & 1 & 1 \\
machines & Our servants, like mere \textbf{machines}, move on their mercenary track without feeling. & 1 & 0 \\
machinery & We have everywhere water power to any desirable extent, suitable for propelling all kinds of \textbf{machinery}. & 0 & 0 \\
\bottomrule
\end{tabular}
\caption{Examples of sentences from the \textit{19thC Machines} dataset with their target expression and corresponding annotations in terms of animacy and humanness.}
\label{tab:19thCexamples}
\end{table*}

\subsection{Datasets summary and discussion}\label{corpus:summary}
Table \ref{tab:jahanVs19thC} summarizes the main differences between the two datasets. \textit{Stories} is larger and more varied in terms of unique target expressions, and has a nearly-perfect inter-annotation agreement (Cohen's $\kappa$ of 0.99).\footnote{Previous work on animacy in Dutch reported a similarly high $\kappa$ agreement score of 0.95 \cite{karsdorp2015animacy}.} \textit{19thC Machines} consists of 593 sentences with 13 unique target expressions, which can be either animate or inanimate, depending on the context.\footnote{Some sentences were discarded because they mentioned a human boiler instead of a boiler engine. Only 13 of the curated machine words appear in the annotations, due to the comparatively lower frequency of the other word forms in the corpus. There are 201 sentences in which the machine has been tagged as animate, of which 81 are also instances of humanness. The \textit{19thC Machines} dataset is publicly available at \href{https://doi.org/10.23636/1215}{https://doi.org/10.23636/1215}.} As discussed, the disagreement was quite high in comparison, indicating that detecting atypical animacy can be a very semantically complex problem, in particular in highly figurative language.

\begin{table*}[h!]
\small
\centering
\begin{tabular}{lccccc}
\toprule
& \#Sentences & IAA & UniqueExpressions & Train/Test & \#Animate \\
\midrule
\textit{Stories} & 18,708 & Cohen's $\kappa$ 0.99 & 5,340 & 13,096/5,612 & 11,666 \\
\textit{19thC Machines} & 593 & Krippendorff's $\alpha$ 0.74 & 13 & 178/415 & 201 \\
\bottomrule
\end{tabular}
\caption{Comparison between the \textit{Stories} and \textit{19thC Machines} animacy datasets.}
\label{tab:jahanVs19thC}
\end{table*}

\section{Language Models}\label{sec:lm}
In our experiments, we used the `BERT base uncased' model and tokenizer as contemporary model,\footnote{\href{https://github.com/google-research/bert}{https://github.com/google-research/bert}.} hereinafter referred to as \textit{BERT-base}. In addition, in order to investigate pattern changes over time, we also fine-tuned \textit{BERT-base} on the \textit{19thC BL Books} dataset, split into four periods (before 1850, between 1850 and 1875, between 1875 and 1890, and between 1890 and 1900), each containing $\approx$1.3B words per period, except for the 1890-1900 time period which had $\approx$940M words.\footnote{While the data distribution for fine-tuning was decided on largely by the number of tokens, these periods also work well in representing distinct cultural eras. For example, the pre-1850 dataset sets apart the first industrial revolution from later developments in Britain. Likewise, 1890-1900 is set off, especially in literary terms, by the emergence of `modernist' sensibilities and the questioning of class and gender hierarchies associated with the term `\textit{fin de si\`{e}cle}'.} The fine-tuning was done in four sequential steps. The \textit{BERT-base} model was first fine-tuned on the oldest time period (i.e., books published before 1850). We then used the resulting language model and further fine-tuned it on the next time period. This procedure of fine-tuning a language model on the subsequent time period was repeated for the other two time periods. For each time period, we preprocessed all books\footnote{We normalized white spaces, removed accents and repeated ``.'' (as they are common in the OCR'd texts), added a white space before and after punctuation signs, and finally split token streams into sentences using the \textit{syntok} library: \href{https://pypi.org/project/syntok/}{https://pypi.org/project/syntok/}.} and tokenized them using the original BERT-base tokenizer as implemented by HuggingFace\footnote{\href{https://github.com/huggingface/transformers}{https://github.com/huggingface/transformers}.} \cite{Wolf2019HuggingFacesTS}. We did not train new tokenizers for each time period. This way, the resulting language models can be compared easily with no further processing or adjustments. The tokenized sentences are then fed to the language model fine-tuning tool in which only the masked language model (MLM) objective is optimized.\footnote{We used a batch size of 5 per GPU and fine-tuned for 1 epoch over the books in each time-period. The choice of batch size was dictated by the available GPU memory (we used 4$\times$ NVIDIA Tesla K80 GPUs in parallel). Similar to the original BERT pre-training procedure, we used the Adam optimization method \cite{kingma2014adam} with learning rate of 1e-4, $\beta_{1} = 0.9$, $\beta_{2} = 0.999$ and $L_{2}$ weight decay of 0.01. In our fine-tunings, we used a linear learning-rate warmup over the first 2,000 steps. A dropout probability of 0.1 was used in all layers.}

We do not aim at modeling animacy change diachronically in this paper. Instead, we treat the different fine-tuned models as four different snapshots of time that we can then compare. Table \ref{tab:ftTimePeriod} shows how our four fine-tuned language models differ in predicting the same masked element in a sentence. While this is a cherry-picked example, it illustrates the importance of having language models that adequately reflect language contemporary to our data.

\begin{table*}[h!]
\centering
\small
\begin{tabular}{p{0.15\linewidth}p{0.75\linewidth}}
\toprule
FT BERT model & Predicted tokens \\
\midrule
$\leq$1850 & \textit{man (5.3291), prisoners (4.9758), men (4.885), book (4.6477), people (4.556), one (4.4271), slave (4.4034), air (4.1329), water (4.1148)} \\
1850--1875 & \textit{men (10.7655), people (9.497), miners (9.249), engine (8.0428), women (8.0126), company (7.7261), machine (7.6021), labourers (7.5987), machines (7.5012)} \\
1875--1890 & \textit{men (10.2048), miners (7.6654), machines (7.4062), people (7.2991), engine (7.232), labourers (7.0957), engines (6.7786), engineers (6.5642), machine (6.4712)} \\
1890--1900 & \textit{mercury (8.0446), machinery (7.4067), machine (7.2903), mine (7.274), mill (7.057), men (7.0257), engine (6.9966), lead (6.9177), miners (6.7764)} \\
\bottomrule
\end{tabular}
\caption{This table illustrates the differences between language models fine-tuned over time: the \textit{Predicted tokens} column contains the tokens predicted as most probable to fill the [MASK] gap in the sentence ``They were told that the [MASK] stopped working'', according to each model. The more recent models clearly predict machine-related words more often than the older ones.}
\label{tab:ftTimePeriod}
\end{table*}

\section{Experiments and results}\label{sec:experiments}

\subsection{Baselines and Evaluation Metrics}
We provide two different types of competitive baselines: a classification approach and a sequence tagging approach. As a reference, we also provide the results obtained by selecting the most frequent class, which is the animate class in the \textit{Stories} dataset and the inanimate class in the \textit{19thC Machines} dataset.

\begin{itemize}
    \item \textbf{Classification approach.} \newcite{karsdorp2015animacy} and \newcite{jahan2018new} treat animacy as a classification problem. We report the performance of three different supervised classifiers trained on examples annotated with a binary label: two SVMs, one using tf-idf and the other using word embeddings as features, and a BERT classifier (from now on \textit{SVM TFIDF}, \textit{SVM WordEmb}, and \textit{BERTClassifier}, respectively).\footnote{Both SVMs use a linear kernel with standard parameters, from the \textit{Scikit-learn} package. We have used the following \textit{Scikit-learn} wrapper for training the BERT classifier, again with default parameters: \href{https://github.com/charles9n/bert-sklearn}{https://github.com/charles9n/bert-sklearn}.} All classifiers are trained both at the target expression level (\textit{targetExp}), and at the context level (i.e.~trained on the target expressions and $n$ words to the left and to the right, where $n$ is 3, as in \newcite{karsdorp2015animacy}), either including the target expression itself (\textit{targetExp + ctxt}) or replacing it with a mask (\textit{maskedExp + ctxt}). For the \textit{Stories} dataset, we used the training set of 13,096 instances to find the optimal hyperparameters and thresholds, which we then applied to the test set. For the \textit{19thC Machines} dataset, due to the small amount of domain-specific data, we trained the classifiers on the concatenation of the \textit{Stories} and \textit{19thC Machines} training sets (resulting in a total of 13,274 instances) and used the \textit{19thC Machines} training set (178 instances) to find the optimal thresholds.
    \item \textbf{Sequential tagging approach}: Moreover, we experimented with LSTMs \cite{hochreiter1997long,akbik-etal-2019-flair} to predict the animacy of target tokens (\textit{SeqModel: LSTM}). In this scenario, we enclosed the target words with `[TARGET]' token --- so the classifier knows which words to tag --- and labeled them as 0 or 1. We assigned `O' to all tokens outside of the `[TARGET]' tokens. We used embeddings extracted from a BERT model as input for a Sequential LSTM tagger. The model is trained and tested on the \textit{Stories} dataset. To assess performance on the \textit{19thC Machines} dataset, we further fine-tuned the \textit{Stories} model on the sentences from the \textit{19thC Machines} training set.
\end{itemize}

\noindent Since our datasets are not always balanced and we want to give equal importance to each class, we report on macro precision and recall, and macro average F-score. For reference, we also provide mean average precision (Map), a popular metric in information retrieval which highlights how well the ranking of the animacy score correlates with the labels.

\subsection{Experimental results}


Table \ref{tab:combinedResults} reports the performance of the different baselines and methods on the \textit{Stories} and \textit{19thC Machines} datasets. The sequence tagging approach (\textit{SeqModel: LSTM}) and classifiers (especially \textit{BERTClassifier}) based on the target expression alone (\textit{targetExp}) are the best performing methods in the \textit{Stories} dataset.
Interestingly, the classifiers' performances become worse when more context is added (\textit{targetExp + ctxt}), and even more so when the target expression is masked (\textit{maskedExp + ctxt}). Our method (\textit{MaskPredict: BERT-base}) does not use the target expression as a feature at all: it relies solely on context. In fact, adding context (one sentence to the left and to the right, \textit{MaskPredict: BERT-base + ctxt}) improves the F-Score from 0.72 to 0.81. This analysis shows that the target expression is a very indicative feature of conventional animacy. And yet, the good performance of our context-based method proves that animacy is not only determined at the level of the target expression, but informed by contextual cues as well.

\begin{table*}[h!]
\centering
\small
\begin{tabular}{l|cccc|cccc}
\toprule
& \multicolumn{4}{|c}{\textit{Stories}}& \multicolumn{4}{|c}{\textit{19thC Machines}}\\
     & Precision & Recall & F-Score & Map & Precision & Recall & F-Score & Map \\
\midrule
Most frequent class & 0.31 & 0.5 & 0.383 & 0.623 & 0.336 & 0.5 & 0.402 & 0.318 \\
\midrule
SVM TFIDF: targetExp & 0.911 & 0.893 & 0.902 & 0.928 & 0.696 & 0.713 & 0.704 & 0.474 \\
SVM WordEmb: targetExp & 0.927 & 0.919 & 0.923 & 0.954 & 0.694 & 0.711 & 0.702 & 0.499 \\
BERTClassifier: targetExp & 0.951 & \textbf{0.948} & 0.949 & \textbf{0.985} & 0.698 & 0.715 & 0.706 & 0.51 \\
\midrule
SVM TFIDF: targetExp + ctxt & 0.734 & 0.739 & 0.737 & 0.859 & 0.688 & 0.71 & 0.699 & 0.651 \\
SVM WordEmb: targetExp + ctxt & 0.758 & 0.742 & 0.75 & 0.876 & 0.728 & 0.531 & 0.614 & 0.481 \\
BERTClassifier: targetExp + ctxt & 0.931 & 0.926 & 0.929 & 0.978 & 0.695 & 0.721 & 0.708 & 0.721 \\
\midrule
SVM TFIDF: maskedExp + ctxt & 0.674 & 0.677 & 0.675 & 0.804 & 0.592 & 0.6 & 0.596 & 0.498 \\
SVM WordEmb: maskedExp + ctxt & 0.674 & 0.678 & 0.676 & 0.809 & 0.518 & 0.52 & 0.519 & 0.339 \\
BERTClassifier: maskedExp + ctxt & 0.855 & 0.852 & 0.854 & 0.951 & 0.687 & 0.696 & 0.692 & 0.603 \\
\midrule
SeqModel: LSTM & \textbf{0.952} & \textbf{0.948} & \textbf{0.95} & 0.949 & 0.697 & 0.719 & 0.708 & 0.482 \\
\midrule
\midrule
MaskPredict: BERT-base & 0.739 & 0.703 & 0.72 & 0.848 & 0.719 & 0.742 & 0.73 & 0.74 \\
MaskPredict: BERT-base +ctxt & 0.839 & 0.774 & 0.806 & 0.892 & 0.758 & \textbf{0.778} & 0.768 & \textbf{0.795} \\
MaskPredict: fit19thBERT +ctxt & -- & -- & -- & -- & 0.758 & 0.775 & 0.766 & 0.777 \\
MaskPredict: early19thBERT +ctxt & -- & -- & -- & -- & \textbf{0.799} & 0.773 & \textbf{0.786} & 0.784 \\
\bottomrule
\end{tabular}
\caption{Evaluation results on the \textit{Stories} and \textit{19thC Machines} dataset.}
\label{tab:combinedResults}
\end{table*}

All baselines perform worse on the \textit{19thC Machines} dataset. As mentioned, in this case we trained the classifiers on the concatenation of the \textit{Stories} and the \textit{19thC Machines} training sets, and optimal classification thresholds were found by maximizing F-Score on the \textit{19thC Machines} training set. The sequential model was trained on the \textit{Stories} training set and fine-tuned on the \textit{19thC Machines} training set. 
Our approach yields consistent performance on both datasets, showing the advantage of its unsupervised context-dependent architecture, as opposed to the other competitive baselines, which heavily rely on the availability of in-domain training data.


We show in Table \ref{tab:combinedResults} the results of using different BERT models in our masking approach: \textit{fit19thCBERT} uses the appropriate fine-tuned BERT model of the period to which each sentence belongs (e.g. a sentence from a book published in 1856 is processed using the BERT model fine-tuned on books published between 1850 and 1875), whereas \textit{early19thCBERT} uses the BERT model fine-tuned on the first half of the nineteenth-century. While the latter provides the best results, the differences are not statistically significant.



\section{Discussion and interpretation}

Researchers across many disciplines 
have long debated the relation between language and the social worlds in which it exists. Studying the linguistic depictions of machines \textit{as if} they were alive raises important questions about the relationship between humans and machines that go beyond language. Animacy and its related concept of agency \cite{yamamoto2006} are important markers of social and political power: when ascribed to non-human actors they indicate the shifting perception of human agency in distinction to that of machines. 
As discussed in section \ref{sec:19thAnnotations}, we consider entities to be animate if they are given attributes and faculties that are characteristic of living entities; they are attributed \textit{humanness} if they are portrayed as sentient and capable of specifically human emotions.\footnote{Example \ref{ex:humanAsMachines} shows sentences in which the machine is animate but lacking humanness. An example of a sentence displaying humanness is: `He bore the movement well [...] and make[s] one wonder why the poor crazy \textbf{machine} is thought worthy of being put together again, with infinite pains, and wonderful science.'} The latter is loosely tied to the idea of an anthropocentric hierarchy in animacy \cite{comrie,croft2002typology,yamamoto1999animacy}, which ranks entities most capable of human perception as the most animate, reflecting notions of agency, closeness to the speaker, or speaker empathy. 

All baselines and methods are clearly worse in predicting humanness than 
animacy.\footnote{Performance for `humanness' is lower overall, with the F-Score of our best-performing method decreasing from 0.79 to 0.63. Due to space limitations, we could not include the evaluation on \textit{humanness} annotations in this paper, but it can be found on our Github repository: \href{https://github.com/Living-with-machines/AtypicalAnimacy}{https://github.com/Living-with-machines/AtypicalAnimacy}.} In the case of the supervised baselines, this is possibly due in part to the very limited amount of annotated data for this concept (81 instances of animacy with humanness, 120 instances of animacy without humanness). Besides, the lower agreement between annotators in detecting humanness (Krippendorff $\alpha$ of 0.50) suggests that the task is more subjective. 
Table \ref{tab:humannessPredictions} reports the top predicted tokens by the pre-1850 BERT model in sentences where animated machines are attributed or negated humanness. This experiment shows how language models can hint a social bias embedding in nineteenth-century discourse.\footnote{Contextualized word embeddings have been used in the past to identify cultural and social biases that permeate language \cite{kurita-etal-2019-measuring}.} While `man' remains the most predicted token replacing machine (and `woman' is not far behind), the appearance of `slave(s)' and `savage' in contexts of negated humanness reflects the nineteenth-century tendency to use these words in discourses that confer diminished human qualities on those people.


\begin{table*}[h!]
\centering
\small
\begin{tabular}{ll}
\toprule
Humanness & Most predicted tokens \\
\midrule
Attributed & \textit{man, woman, engine, person, child, dog, fellow, boy, machine, men} \\
Negated & \textit{man, men, soldier, horse, savage, coward, slave, woman, creatures, slaves} \\
\bottomrule
\end{tabular}
\caption{Most predicted tokens in sentences in which the machine is attributed humanness or lack thereof, according to the pre-1850 BERT model.}
\label{tab:humannessPredictions}
\end{table*}

\noindent Language is not static, it evolves continuously \cite{hamilton-etal-2016-diachronic,kutuzov}. Social and technological changes are paralleled by changes in the language used to describe them. In the nineteenth century, who or what was performing work was changing dramatically, with different groups of people entering and exiting the (free and unfree) labor pool and becoming key parts of the workforce at different times. 

\begin{table*}[h!]
\centering
\small
\begin{tabular}{p{0.45\linewidth}p{0.22\linewidth}p{0.22\linewidth}}
\toprule
Sentence & $\leq$1850 BERT & Contemporary BERT  \\
\midrule
Our sewing \textbf{machines} stood near the wall where grated windows admitted sunshine, and their hymn to Labour was the only sound that broke the brooding silence. & \textit{girls, women, men, boys, children, machines, parties, females, mothers, hands} & \textit{machines, tables, stalls, machine, rooms, sheds, carts, girls, boxes, women} \\
\bottomrule
\end{tabular}
\caption{Top predictions for the word \textit{`machines'}, using the pre-1850 and contemporary BERT models.}
\label{tab:sewing}
\end{table*}

\noindent Such changes are often implicitly captured by a language model. Table \ref{tab:sewing} gives a concrete example, related to the activity of `sewing', which was originally performed by humans (often women and children) but increasingly mechanized. This shift is reflected in the difference between the predictions of the pre-1850 and the contemporary language model. In future work, we will explore how language models encode biases and attitudes, and therefore can be used to scrutinize subtle but critical shifts in the understanding of technology and mechanization. This is relevant not only to nineteenth-century discourses of industrialization, but also to contemporary discussion of the impact of technology in our society.

\section{Conclusion and further work}
We have introduced a new method for animacy detection based on contextualized word embeddings, which efficiently  handles atypical animacy. Our case study explores how machines were portrayed in nineteenth-century texts and is motivated by the ubiquitous trope of the living machine; both in the historical discourse of industrialization, and also in today's discussion of AI and robotics, prefigured by Alan Turing's famous provocation: `Can machines think?' \cite{turing1950computing}. This work opens many avenues for future research. We intend to explore time-sensitive strategies to derive an animacy value from BERT's predictions by inspecting the embedding space; study the contextual cues which grant animacy (and how these relate to the neighboring concepts of humanness  and agency); and explore the extent to which such atypicalities are conveyed through figurative language. Finally, we will apply all of the above in addressing the historical questions raised in this paper.

\bibliographystyle{acl}
\bibliography{references}

\end{document}